\documentclass[letterpaper, 10 pt, conference]{ieeeconf}  

\IEEEoverridecommandlockouts                              

\usepackage{graphicx} 
\usepackage{amsmath} 
\usepackage{amssymb}  
\usepackage{setspace}
\usepackage{algpseudocode}
\usepackage{xcolor}
\usepackage{lipsum}
\usepackage{booktabs}
\usepackage{adjustbox}
\usepackage{mathtools}
\usepackage{float}
\usepackage{xfrac}
\usepackage{hyperref}

\definecolor{ForestGreen}{rgb}{0.133, 0.545, 0.133}
\usepackage[ruled,vlined,linesnumbered]{algorithm2e}

\SetCommentSty{mycommfont}
\title{\LARGE \bf
Achieving Autonomous Cloth Manipulation with Optimal Control via Differentiable Physics-Aware Regularization and Safety Constraints 
}

\author{
Yutong Zhang$^{\dagger, 1}$, Fei Liu$^{\dagger, 1}$, Xiao Liang$^1$, and Michael Yip$^1$ \IEEEmembership{Senior Member, IEEE}
\thanks{$^{1}$ Department of Electrical and Computer Engineering, University of California San Diego,
La Jolla, CA 92093 USA. {\tt\small\{yuz049, f4liu, x5liang, yip\}@ucsd.edu}, $^\dagger$ represents equal contributions.}%
}

\begin{document}

\maketitle
\thispagestyle{empty}
\pagestyle{empty}

\begin{abstract}
Cloth manipulation is a category of deformable object manipulation of great interest to the robotics community, from applications of automated laundry-folding and home organizing and cleaning to textiles and flexible manufacturing. Despite the desire for automated cloth manipulation, the thin-shell dynamics and under-actuation nature of cloth present significant challenges for robots to effectively interact with them. Many recent works omit explicit modeling in favor of learning-based methods that may yield control policies directly. However, these methods require large training sets that must be collected and curated. In this regard, we create a framework for differentiable modeling of cloth dynamics leveraging an Extended Position-based Dynamics (XPBD) algorithm. Together with the desired control objective, physics-aware regularization terms are designed for better results, including trajectory smoothness and elastic potential energy. In addition, safety constraints, such as avoiding obstacles, can be specified using signed distance functions (SDFs). We formulate the cloth manipulation task with safety constraints as a constrained optimization problem, which can be effectively solved by mainstream gradient-based optimizers thanks to the end-to-end differentiability of our framework. Finally, we assess the proposed framework for manipulation tasks with various safety thresholds and demonstrate the feasibility of result trajectories on a surgical robot. The effects of the regularization terms are analyzed in an additional ablation study.
\end{abstract}



\section{INTRODUCTION}

Cloth manipulation has numerous applications both in housework and manufacturing for robotics, but it remains challenging due to the material's deformable, flexible, and dynamic nature. The manipulation of cloth can be seen as an overall trajectory planning process with a focus on changing the shape configuration while maintaining geometrical or topological properties. Unlike rigid objects, cloth is difficult to manipulate due to the under-actuated control with infinite degrees of freedom for shape deformation. Meanwhile, slight disturbances on cloth can result in significantly dynamic behaviors, such as bending, crumpling, and self-occlusion \cite{Solvi_2023}. Study into cloth manipulation in the robotics community has covered various aspects, including visual representation \cite{Xiao_Ma_2022}, latent-space modeling \cite{Wilson_2021}, cloth grasping primitive \cite{Borras_2020_TRO}, sequential multi-step control \cite{Kai_2023_RAL}, imitation learning \cite{Chen_2023_CVPR} and reinforcement learning \cite{Jangir_2020_ICRA}.

Trajectories play a pivotal role in addressing multiple safety-critical tasks. In contrast to rigid objects, cloth dynamics exhibit diverse responses influenced by their interactions with the environment, primarily through deformation. Consequently, incorporating considerations of reliability and safety into cloth control poses a significant challenge for modeling and planning. Even with the integration of a high-fidelity deformable model for cloth with constraints, the task of trajectory optimization remains challenging \cite{Jihong_2022, Hang_Yin_2021, Doumanoglou_2016}.

\begin{figure}[!t]
    \centering
    \includegraphics[width=\linewidth]{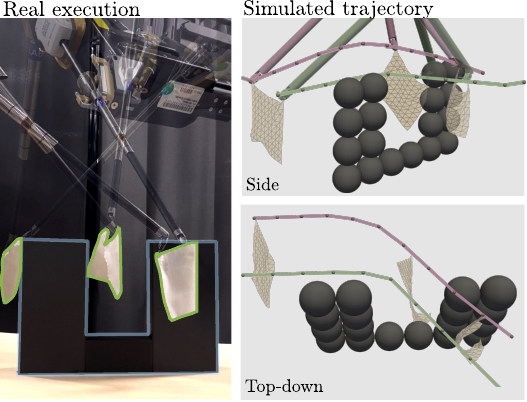}
    \caption{Execution of an optimized physics-aware and safe cloth manipulation trajectory in real work on a dVRK system (left) and in a XPBD simulation with robot kinematics (right). The obstacles are represented with spheres in the simulator. We show that we can make the cloth follow trajectories through constrained environments while achieving goal configurations under dynamic motion.}
    \label{fig:cover_photo}
    \vspace{-1.5em}
\end{figure}
Therefore, our objective is to build a physics-based approach for manipulating cloth that can generate control trajectories while taking into account both the dynamics of cloth and regulations of trajectories for safety. The ideal control sequence should ensure the cloth to a target position through a smooth trajectory without overstretching the cloth or breaking safety constraints.



\subsection{Contributions}
In recent years, simulations of deformable objects have been increasingly developed in the field of Computer Graphics. By simply defining various geometrical constraints, position-based dynamics (PBD) has proven to be a promising method for fast online simulation of deformable objects \cite{Bender_2017, Fei_2023_RAL, Fei_2021_ICRA, Jingbin_2021_ICRA, Yunhai_2020}. Based on the extended position-based dynamics (XPBD) algorithm \cite{Macklin_2016_XPBD}, we propose a framework for cloth manipulation that regularizes physics-aware metrics and complies with safety constraints. To this end, we present the following novel contributions :
\begin{itemize}
    \item We leverage the full differentiability of the developed XPBD simulation to power a gradient-based trajectory optimization framework for cloth manipulation.
    \item Physics-aware regularization terms are introduced and integrated into our framework to prevent undesired behaviors, such as over-stretching and wrinkling.
    \item Safety constraints such as obstacle avoidance requirements are formulated as a constraint function that can be incorporated into the trajectory optimization framework.
    \item We evaluate the framework with non-trivial cloth manipulation tasks. Different metrics are extensively analyzed and
    the trajectory is executed by the dVRK robot to verify its compatibility with real robot kinematics.
\end{itemize}

\section{Related Works}
\subsection{Cloth Dynamics Modeling}

Research in cloth dynamics, explored in computer graphics and robotics, encompasses various modeling techniques, from mass-spring systems to FEM-based continuum mechanics \cite{James_2011, Hang_Yin_2021}. These models have informed cloth manipulation strategies for motion controllers \cite{Yunfei_2016, McConachie_2020_IJRR}, but can be challenged by complex environments and local minima.

Recently, deep learning has driven interest in data-driven approaches. One group employs neural networks to create cloth dynamics models, applying latent space control \cite{Xiao_Ma_2022, Wilson_2021, Cheng_Chi_2022_RSS}. Another group uses model-free methods, such as deep imitation learning \cite{Chen_2023_CVPR, Seita_2020_IROS, Jia_2019_RAL}, dynamic movement primitives \cite{Fan_Zhang_2022_ScienceRobotics, Joshi_2017}, and reinforcement learning \cite{Jangir_2020_ICRA, Xingyu_2020, Yilin_2020_RSS}, often based on trial-and-error in simulation. However, these methods may face challenges with implicit models and costly real robot training with defined reward functions.

\subsection{Safe Manipulation}



In recent years, robotics has focused on safety planning and control, driven by methods like control barrier functions (CBFs) \cite{Wei_Xiao_2023_TRO, Singletary_2022_RAL, Wang_Li_2017_TRO}. However, these methods mainly apply to rigid robots, addressing safety constraints like contact forces and joint limits. Research spans legged robot locomotion \cite{Villarreal_2020_ICRA, Grandia_2021_ICRA} and robotic manipulators \cite{Nubert_2020_RAL, Yilun_2023}.

Conversely, cloth manipulation, particularly with a safety focus, has seen limited exploration. Recent work primarily adopts learning-based approaches for safe manipulation policies \cite{Joshi_2019, Alexander_2018} and learning to ensure safe interactions between cloth and its environment \cite{Dragusanu_2022_RAL, Mitrano_2021_ScienceR}. Notably, Erickson et al. \cite{Erickson_ICRA_2018} demonstrated using learned models for safe model predictive control. However, these methods rely on substantial training data, which could benefit from incorporating physics simulations.

\section{Methodology}
\subsection{Problem Formulation}
The state of the cloth is given by $\mathbf{x} \in \mathbb{R}^{N \times 3}$, where $N$ is the number of particles in the cloth mesh.
At each time step $t$, $\mathbf{x}^{(t)}$ can be computed from previous state $\mathbf{x}^{(t-1)}$ by applying control $\mathbf{u}^{(t-1)}$.
The control sequence $\mathbf{U}$ is characterized as a list of displacement vectors associated with specific control points.
Given control sequence $\mathbf{U} = [\mathbf{u}^{(0)}, \mathbf{u}^{(2)}, \dots, \mathbf{u}^{(T-1)}]^\top$ and initial state $\mathbf{x}^{init}$, we can obtain a sequence of intermediate states $\mathbf{X} = [\mathbf{x}^{(1)}, \mathbf{x}^{(2)}, \dots, \mathbf{x}^{(T)}]^\top$.
Then, the cloth manipulation task can be formulated as a trajectory optimization problem, given by:
\begin{equation}\small
\begin{aligned}
\label{eq:optimization_formulation}
    \arg\min_{\mathbf{U}}& && \mathcal{L}(\mathbf{X}, \mathbf{U}) \\
    \text{s.t.}& && \mathbf{x}^{(t)} = \mathcal{XPBD}\left(\mathbf{x}^{(t-1)}, \mathbf{u}^{(t-1)}\right)\\
    & && \mathbf{x^{(0)}} = x^{init}\\
    & && \mathcal{C}(\mathbf{X}) \geq 0
\end{aligned}
\end{equation}
where $\mathcal{L}$ is the objective function that includes error and regularization terms, $\mathcal{XPBD}$ represents the extended position-based dynamics algorithm \cite{Macklin_2016_XPBD}, and $\mathcal{C}$ specifies safety constraints on all intermediate states of the cloth.

\subsection{XPBD Simulation}
PBD \cite{Bender_2017} is a physics simulation algorithm based on geometric constraints and positional updates that models deformable objects as a discrete system with particles. 
Compared with traditional force-based methods, PBD is more stable and controllable and converges orders-of-magnitude faster than force-based methods. These benefits make PBD widely used in real-time interactive scenarios. XPBD \cite{Macklin_2016_XPBD} is an extension to the original PBD algorithm to address the issue of iteration and time step-dependent stiffness. It also builds the connection between geometric constraints and elastic potential energy. Later, we will use this physical interpretation to formulate a regularization term that constrains undesired cloth deformation.
\begin{figure}[!b]
    \centering
    \includegraphics[width=0.95\linewidth]{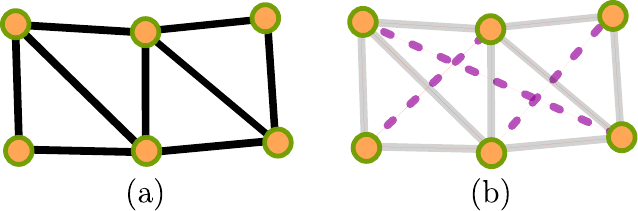} 
    \caption{Visualization of distance and bending constraints in particle-mesh structure.
    $(a)$ A triangular mesh with particles and edges. 
    $(b)$ Distance constraints (gray solid line) and bending constraints (purple dashed line). 
    }
    \label{fig:constraint}
    \vspace{-1.5em}
\end{figure}

In this work, we implement a \textit{quasi-static} XPBD simulation to simulate the state of the cloth under a slow-moving control sequence. 
In a quasi-static process, the state of the cloth evolves from one state to another infinitesimally slowly, ensuring that it always remains in equilibrium.
From this assumption, velocity terms can be ignored. We use distance and bending constraints to model the thin-shell deformation of cloth, as shown in Fig. \ref{fig:constraint}.
\begin{itemize}
    \item \textbf{Distance Constraint} $\mathbf{C}_{dist}(\mathbf{x_i}, \mathbf{x_j})$ is defined between each pair of connected particles in each triangle.
    \item \textbf{Bending Constraint} $\mathbf{C}_{bend}(\mathbf{x_i}, \mathbf{x_j})$ is defined between each pair of non-neighboring particles in each pair of adjacent triangles. 
\end{itemize}
The shared constraint function is identically defined as:
\begin{equation}
\label{eq:dis_bend_constraint}
    \mathbf{C}_{dist/bend}(\mathbf{x}_i, \mathbf{x}_j) = \|\mathbf{x}_i - \mathbf{x}_j\| - \mathbf{d}^0_{ij}
\end{equation}
where $\mathbf{d}^0_{ij}$ is the initial distance between $\mathbf{x}_i$ and $\mathbf{x}_j$. More details of these constraint functions and their gradients can be found in \cite{Bender_2017}.

\begin{figure}[!t]
    \centering  
    \includegraphics[width=0.95\linewidth]{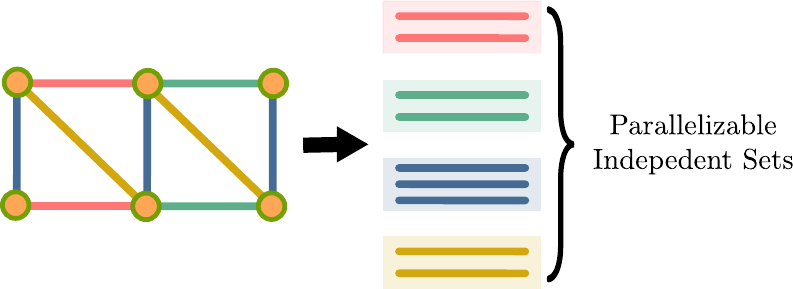}
    \caption{An example of constructing independent sets of constraints. In this example mesh, distance constraints are grouped into four independent sets labeled with different colors. In each set, there is no particle that is shared by any constraint. Therefore, the constraint projection in each set can be done in parallel to speed up the simulation.}
    \label{fig:independent_set}
    \vspace{-1.5em}
\end{figure}

We adopt the Gauss-Seidel style solver~\cite{Muller_2007_PBD}, where the result from one constraint projection is used in the next constraint projection, as it exhibits good convergence and stability. To parallelize the constraint projection steps while avoiding data race conditions, we separate constraints into independent sets where there is no particle shared by any constraint, as shown in Fig. \ref{fig:independent_set}. We can then efficiently project each set of constraints in parallel with broadcast syntax.

\subsection{Objective Function and Safety Constraints}

We define the terminal objective of the trajectory optimization problem as positional alignments between a subset of simulation particles $\mathbf{\bar{x}} \subset \mathbf{x}^{(T)}$ and some user-defined target position $\mathbf{x}_{target}$.
The \textbf{target error} is the $L_2$ distance between the subset of particles and the target position.
\begin{equation}
    \label{eq:target_error}
    \mathcal{G}(\mathbf{x}^{(T)}, \mathbf{x}_{target}) = \|\mathbf{x}_{target} - \mathbf{\bar{x}}\|
\end{equation}

By only considering the terminal alignment of particles, the framework may generate
irregular trajectories inducing excessive deformation on intermediate states of the cloth.
To ensure the smoothness of the trajectory $\mathbf{U}$, the difference between two neighboring control vectors should be limited.
Therefore, we regularize the \textbf{trajectory irregularity} as the sum of the backward difference of the control sequence in discrete time steps.
\begin{equation}\small
    \label{eq:traj_irregularity}
    \mathcal{T}(\mathbf{U}) = \sum\nolimits_{t=1}^{T} \|\mathbf{u}_{i} - \mathbf{u}_{i-1}\|
\end{equation}

We emphasize that an optimal control sequence is aware of the deformation physics of the cloth, preventing undesirable overstretching and bending effects. This can be accomplished by regularizing the potential energy of the cloth over time. In our XPBD simulation, the total potential energy is the sum of the energy of every geometric constraint.
Let $\mathbf{C}(\mathbf{\mathbf{x}^{(t)}})=[C_1(\mathbf{\mathbf{\mathbf{x}^{(t)}}}), C_2(\mathbf{\mathbf{\mathbf{x}^{(t)}}}), \dots, C_m(\mathbf{\mathbf{x}^{(t)}})]^\top$ be a set of geometric constraints, and let $\mathbf{K}=[k_1, k_2, \dots, k_m]^T$ be their associated stiffness parameters.
The \textbf{potential energy} of the cloth through the control trajectory can be expressed as:
\begin{equation}\small
    \label{eq:potential_energy}
    \begin{split}
            \mathcal{E}(\mathbf{X}) = \sum\nolimits_{t=1}^T \large \sfrac{1}{2}\  \mathbf{C}(\mathbf{x}^{(t)})^\top \mathtt{diag}(\mathbf{K}) \mathbf{C}(\mathbf{x}^{(t)})
    \end{split}
\end{equation}

The objective function is then the combination of target error,
trajectory irregularity and potential energy regularization:
\begin{equation}
\begin{split}
    \mathcal{L}(\mathbf{X}, \mathbf{U}) = \mathcal{G}(\mathbf{x}^{(T)}, \mathbf{x}_{target}) + \alpha  \mathcal{T}(\mathbf{U})
    + \beta \mathcal{E}(\mathbf{X})
\end{split}
\end{equation}
Here, $\alpha$ and $\beta$ are scaling coefficients on two regularization terms.
Later we will study the effects of the regularization terms in an ablation study.


Safety constraints are essential in many manipulation tasks.
Our formulation \ref{eq:optimization_formulation} allows us to incorporate various safety constraints, such as robot joint limit, contact forces, and obstacle avoidance, into the differentiable simulation framework. For simplicity, we demonstrate its ability in an obstacle avoidance scenario.
The distance to the obstacle can be specified by an SDF $\mathcal{SDF}$ where a positive value means outside and a negative value means inside. 
To simplify the computation of the SDF, we approximate different obstacle shapes with numerous spheres.
For a sphere $\mathbf{s}$ centered at point $\mathbf{c}$ with radius $r$, the SDF can be written as:
\begin{equation}
    \mathcal{SDF}_{\mathbf{s}}(\mathbf{x}) = \|\mathbf{x} - \mathbf{c}\| - r
\end{equation}
The SDF of the entire obstacle is then given by:
\begin{equation}
    \mathcal{SDF}(\mathbf{x}) = \min \{ \mathcal{SDF}_{\mathbf{s}}(\mathbf{x}) \text{ for } \mathbf{s} \in \mathbf{S}\}
\end{equation}
where $\mathbf{S}$ is the set of spheres making up the obstacle. This approximation is highly popular in robotic planning applications and in solving for feasible inverse-kinematic solutions in constrained environments. Furthermore, the fact that it is analytically and efficiently evaluated is beneficial to the optimization procedure. We refer readers to Diffco~\cite{zhi2022diffco} for an alterative still-differentiable SDF formulation that goes beyond spherical approximation for generality and is still computationally efficient for trajectory optimization.

\begin{figure}[!b]
    \centering  
    \includegraphics[width=0.95\linewidth]{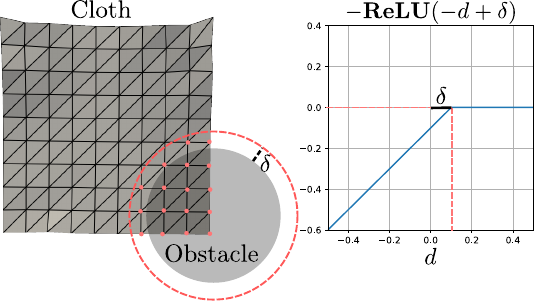}
    \caption{Illustration of the collision constraint.
    On the left, the cloth penetrates the gray obstacle.
    The red dashed circle indicates the range of collision avoidance shifted by the constant $\delta$.
    All the particles result in negative constraint values are marked by the red dots.
    On the right, we plot the function $-\mathbf{ReLU}(-d + \delta)$.
    It is clear that the function value is negative for all $d < \delta$,
    which is up to our control.}
    \vspace{-1.5em}
    \label{fig:collision_function}
\end{figure}
We can record the signed distance $d_i^{(t)} = \mathcal{SDF}(\mathbf{x}_i^{(t)})$ of each particle at each time step.
Then, a constant $\delta > 0$ is subtracted from all signed distances,
which gives us flexible control of how far we want to avoid the obstacle.
To collect all values violating the safety constraint, we modify the widely used Rectified Linear Unit ($\mathbf{ReLU}$) function.
Considering function $-\mathbf{ReLU}(-d + \delta)$, the value will be $d - \delta$ when $d$ is smaller than $\delta$, $0$ when $d$ is no less than $\delta$.
We can then express our obstacle avoidance safety constraint as the sum of all the particles along the control trajectory.
\begin{equation}\small
    \mathcal{C}(\mathbf{X}) = \sum\nolimits_{t=1}^{T} \sum\nolimits_{i=1}^{N} -\textbf{ReLU}(-d_i^{(t)} + \delta)
\end{equation}
This constraint function would be non-negative only when signed distances of all particles are no less than threshold $\delta$.
Fig \ref{fig:collision_function} visualizes an example collision and plots the behavior of the modified $\mathbf{ReLU}$ function we use in the constraint.

\begin{algorithm}[!t]
\setstretch{1.0}
\footnotesize
    \caption{Differentiable $\mathcal{XPBD}$ Cloth Manipulation Framework}
    \label{alg:diff_XPBD}
    \SetKwInOut{Input}{Input}
    \SetKwInOut{Output}{Output}

    \Input{Initial state $\mathbf{x}^{(0)}$,initial control sequence $\mathbf{U}$, distance stiffness $\mathbf{k}_{dis}$, bending stiffness $\mathbf{k}_{bend}$}
    \Output{Optimized control sequence $\mathbf{U}^*$}
        \For{$t$ in $1...T$}{

            $\mathbf{x}^{(t)} \leftarrow \mathbf{x}^{(t-1)} + \frac{1}{2} \mathbf{g} \Delta{t}^2 $

            $\mathbf{x}^{(t)} \leftarrow \mathtt{applyControl}\left(\mathbf{x}^{(t)}, \mathbf{u}^{(t-1)}\right)$

            \tcp{Constraints solving loop}
            \While{iter $<$ iteration}{
                \tcp{Solve distance and bending (Eq. \ref{eq:dis_bend_constraint})}
                $\Delta{\mathbf{x}_{dis}} \leftarrow \mathtt{solveDistance}\left(\mathbf{C}_{dist}\left(\mathbf{x}^{(t+1)}\right), \mathbf{k}_{dis}\right)$

                $\Delta{\mathbf{x}_{bend}} \leftarrow \mathtt{solveBending}\left(\mathbf{C}_{bend}\left(\mathbf{x}^{(t+1)}\right), \mathbf{k}_{bend}\right)$ 
    
                \tcp{Update cloth state}
                $\mathbf{x}^{(t)} \leftarrow \mathbf{x}^{(t)} + \Delta{\mathbf{x}_{dis}} + \Delta{\mathbf{x}_{bend}}$
            }
            \tcp{Compute time step gradient}
            $ \nabla\mathcal{XPBD} \leftarrow \displaystyle \large \sfrac{\partial \mathbf{x}^{(t)} }{\partial \mathbf{u}^{(t-1)}}   $ 
        }

        $\mathbf{X} \leftarrow  \mathbf{x}^{(0)}, \cdots, \mathbf{x}^{(T)}$

        \tcp{Obtain the loss function}
        $\mathcal{L} \leftarrow \mathcal{L} \left(\mathbf{X}, \mathbf{U} \right)$

        \tcp{Obtain the safety constraint function}
        $\mathcal{C} \leftarrow \mathcal{C}\left(\mathbf{X}\right)$
        
        \tcp{Compute optimized control}
	$\mathbf{U}^* \leftarrow  \mathtt{optimizer}\left(\mathcal{L}, \frac{d \mathcal{L}}{d \mathbf{U}}, \mathcal{C}\right)$
\end{algorithm}
\subsection{Differentiation Framework}
To find the control sequence through optimization,
it is necessary to compute the gradient of the objective function with respect to $\mathbf{U}$.
Given objective function function $\mathcal{L}(\mathbf{X}, \mathbf{U})$, by chain rule we have:
\begin{equation}
    \frac{\mathrm{d} \mathcal{L}}{\mathrm{d} \mathbf{U}} =
    \frac{\partial \mathcal{L}}{\partial \mathbf{U}} +
    \frac{\partial \mathcal{L}}{\partial \mathbf{X}}
    \frac{\mathrm{d} \mathbf{X}}{\mathrm{d} \mathbf{U}}
\end{equation}

We can perform gradient-based optimization using Newton or quasi-Newton methods \cite{SciPy_NMeth_2020} \cite{Feinman_2021_minimize}. It should be noticed that the challenge is to evaluate the total derivative $\frac{\mathrm{d} \mathbf{X}}{\mathrm{d} \mathbf{U}}$. According to the XPBD system in Eq. \ref{eq:optimization_formulation}, each Jacobian element in the total derivative can be expressed by:
\begin{equation}
    \left[ \frac{\mathrm{d} \mathbf{X}}{\mathrm{d} \mathbf{U}} \right]^{(p, q)} = \frac{\mathrm{d} \mathbf{x}^{(p)} }{\mathrm{d} \mathbf{u}^{(q)}} 
\end{equation}
where
\begin{equation}\small
    \frac{\mathrm{d} \mathbf{x}^{(p)} }{\mathrm{d} \mathbf{u}^{(q)}} = \left\{ 
    \begin{aligned}
    & \frac{\partial \mathbf{x}^{(p)} }{\partial \mathbf{x}^{(p-1)}} \frac{\mathrm{d} \mathbf{x}^{(p-1)} }{\mathrm{d} \mathbf{u}^{(q)}}, ~~~~~\quad\quad q<p-1 \\[1ex]
    & \nabla_{\mathbf{u}}\mathcal{XPBD}\mid_{(\mathbf{x}^{(p-1)}, \mathbf{u}^{(p-1)})}, q = p-1 \\[1ex]
    & 0, \qquad \qquad \qquad \qquad \qquad \text{otherwise}
    \end{aligned} 
    \right.
\end{equation}
Therefore, it is clear that all we need to do is to compute $\nabla_{\mathbf{u}}\mathcal{XPBD}$ iteratively at each time step.

We detailed our differentiable XPBD simulation in Algorithm \ref{alg:diff_XPBD}. It is implemented as a PyTorch differentiation layer by subclassing the {\tt torch.autograd.Function}. This enables us to back-propagate from any state $\mathbf{x}^{(p)}$ and compute its derivative with respect to past control input $\mathbf{u}^{(q)}$.
The derivatives regarding objective function $\frac{\partial \mathcal{L}}{\partial \mathbf{U}}$ and $\frac{\partial \mathcal{L}}{\partial \mathbf{X}}$ are also calculated using auto-differentiation operations, so the entire computation process is end-to-end differentiable.

\begin{figure*}[!t]
    \centering
    \includegraphics[width=0.95\linewidth]{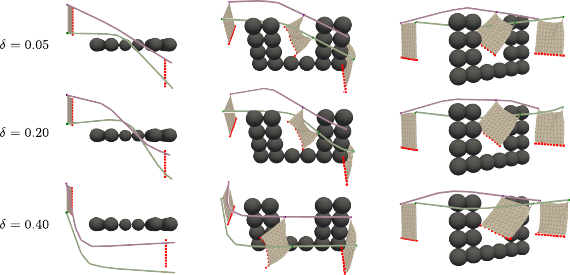}
    \caption{Visualizations of optimized manipulation trajectories with different safety thresholds $\delta$.
    Each row illustrates the trajectory with a safety threshold from different viewing angles.
    Red points denote the target position and the involved particles on the cloth.
    The purple and green lines are the paths of the two control points under the manipulation trajectory.
    The leftmost column only plots the initial state of the cloth.
    The right two columns show the initial state, the terminal state, and one intermediate state of the cloth to
    visualize the whole dynamics under the manipulation trajectory.}
    \label{fig:delta_result}
    \vspace{-1.5em}
\end{figure*}
\section{EXPERIMENTS \& RESULTS}
\subsection{Experiment Setup}
We evaluate the performance of our proposed differentiable cloth manipulation framework on a task with a U-shaped obstacle.
The cloth is discretized into a $10 \times 10$ square mesh.
Distance and bending constraints are defined between these $100$ particles as discussed in section III-B.
Two control points are defined on the top two corners of the cloth.
The control sequence is of length $10$, where each control input is a $2 \times 3$ tensor
specifying displacements on the two control points respectively.
At each time step, the XPBD algorithm solves for $100$ iterations to produce the next state.
The objective function we use is
$\mathcal{L} = \mathcal{G} + \alpha \mathcal{T} + \beta \mathcal{E}$, where $\alpha = 1$ and $\beta=1e-4$.
The safety constraint is to avoid the U-shaped obstacle made up of $18$ uniform spheres.
We can analytically express the SDF of the obstacle,
which is then used to evaluate constraint function $\mathcal{C}$.
We run experiments with a series of safety thresholds $\delta$ and analyze how their values influence the optimized trajectories.

To verify the feasibility of the proposed framework in a real environment,
we obtain the optimized trajectory from the experiment described above and recreate a real scene following the simulation setup.
We utilize the da Vinci Research Kits (dVRK) \cite{dvrk} to manipulate a piece of cloth while avoiding a real 3D-printed U-shaped obstacle. 
The real obstacle has the same shape as the previous experiment with a dimension of $3\times 12 \times 18$ cm, whereas the cloth has a square shape with a size of $6\times 6$ cm.
Finally, we solve for inverse kinematics to generate dense way-points along the discrete trajectory we obtain,
then let the robotic grippers follow the way-points.
While this workflow assumes a pre-known 3D environment for optimization,
it is possible to obtain this information with an RGB-D or stereo-camera setup.

\subsection{Safety Constraints Experiment}
In this experiment, we test our framework with the obstacle avoidance safety constraint.
The cloth is initialized on one side of the U-shaped obstacle and the goal is to align the bottom edge of the cloth to a horizontal line
on the other side of the obstacle.
We compare how various safety thresholds result in different control trajectories.

In Fig \ref{fig:delta_result}, we show how the cloth is moved to the target position under manipulation trajectories with different safety thresholds.
When $\delta = 0.05$, the resulting trajectory moves the cloth diagonally through the opening of the U-shaped obstacle.
When $\delta = 0.20$, the resulting trajectory rotates the cloth more to pass through the opening.
From the top view, we can see that the paths of the two control points almost align with each other at the opening.
This manipulation behavior helps the cloth to maintain a positive distance from the obstacle as specified by the larger safety threshold.
If $\delta$ is further increased to $0.40$, we can observe that the trajectory no longer passes through the U-shaped opening.
Instead, our framework produces a conservative manipulation trajectory where the cloth is moved to the target position by going around the obstacle.

\begin{table}[t!]
\setlength\tabcolsep{1.5em}
\centering
\caption{Resulting Metrics on Different Safety Thresholds }
\begin{adjustbox}{width=0.45\textwidth}
    \begin{tabular}{c|cccc}
        \toprule
         Metrics & $\delta = 0.05$ & $\delta = 0.20$ & $\delta = 0.40$\\
        \midrule
        $\mathcal{G}$ & $0.23$ & $0.27$ & $0.41$\\
        $\mathcal{T}$ & $0.57$ & $1.33$ & $1.30$\\
        $\mathcal{E}$ & $1.65$ & $1.56$ & $1.34$\\
        $\mathcal{C}$ & $-4 \times 10^{-7}$ & $0.00$ & $-6 \times 10^{-7}$\\
        $\min{\mathcal\{SDF\}}$  & $0.05$ & $0.20$ & $0.40$\\
        \bottomrule
    \end{tabular}
\end{adjustbox}
\label{tbl:cost_table}
\vspace{-1.5em}
\end{table}

In Table \ref{tbl:cost_table}, for each safety threshold, we list values of the objective function, constraint function, and the minimum distance from the cloth to the obstacle. Values of the two regularization terms are scaled as mentioned in the setup.
For all three safety thresholds, our framework produces trajectories satisfying the safety constraint $\mathcal{C} \geq 0$ within marginal tolerance.
The minimum distances also agree with the safety threshold as expected.
Examining the target errors, we observe that they fall within reasonable ranges for all trajectories. The trajectory with $\delta = 0.05$ manages to
converge to the lowest target error primarily because its safety constraint is the least restrictive. For same reason, it also has the smallest trajectory irregularity, while the other two trajectories
must undergo more drastic directional changes to avoid obstacles.
It is worth noting that the trajectory with larger safety thresholds also has less total potential energy. This suggests that trajectories optimized with higher safety thresholds potentially prioritize maintaining the cloth's stability and minimizing stretching in order to constrain its motion within the broader safety boundaries.

\begin{figure}[!t]
    \centering
    \includegraphics[width=\linewidth]{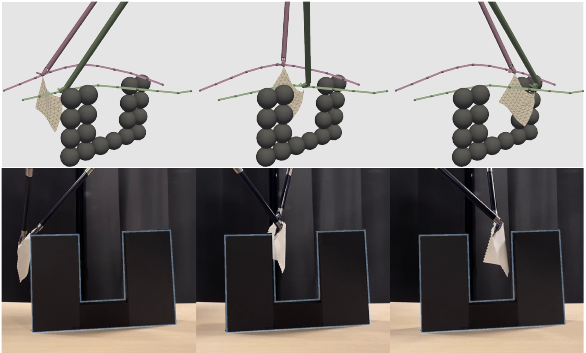}
    \caption{Visualizations of the trajectory in inverse kinematics simulation and on real robot.
    Two robot arms and corresponding control paths are colored by purple and green respectively.
    The obstacle in real experiment is highlighted in blue.}
    \label{fig:traj_in_sim}
    \vspace{-1.5em}
\end{figure}

\begin{figure*}[!t]
    \centering
    \includegraphics[width=1\linewidth]{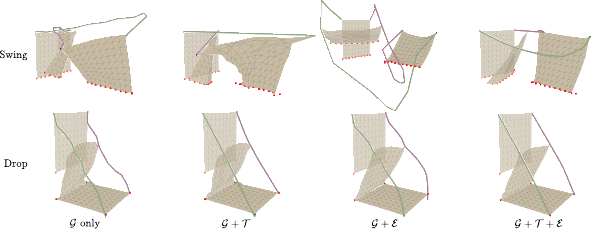}
    \vspace{-1.5em}
    \caption{Visualizations of optimized cloth manipulation trajectories with different regularization
    strategies. The first row shows the results of the swing task. The second row shows
    the results of the drop task. For both tasks, from left to right we optimize for target
    loss $\mathcal{G}$ only, target loss $\mathcal{G}$ with trajectory regularization
    $\mathcal{T}$, target loss $\mathcal{G}$ with energy regularization $\mathcal{E}$
    and target loss $\mathcal{G}$ with both regularization terms $\mathcal{T}$ and $\mathcal{E}$.
    Both $\mathcal{T}$ and $\mathcal{E}$ terms are crucial for finding an efficient and physics-aware control trajectory.}
    \label{fig:qualitative_result}
    \vspace{-1em}
\end{figure*}
In Fig \ref{fig:traj_in_sim}, we fed the trajectory into an inverse kinematics solver and let
the dVRK robot follow the trajectory.
It demonstrates that the trajectory is feasible for the real robot and satisfies the obstacle avoidance safety constraint.
Due to the calibration accuracy, the real experiment does not recreate the control trajectory perfectly.
We can mitigate the calibration error by adjusting the safety threshold to constrain the movement of the cloth. Furthermore, additional techniques (outside the scope of this paper) are available for addressing this calibration gap~\cite{Fei_2021_ICRA}.

\subsection{Ablation Study on Objective function}
We conduct an ablation study on different forms of the proposed objective function. Specifically, we focus on different regularization terms' impact on the quality of optimized trajectories. To facilitate this investigation, we simplify the trajectory optimization problem by isolating the influence of safety constraints, converting it into an unconstrained optimization problem. We introduce two supplemental cloth manipulation tasks for this study. They are (1) Swing: swinging the cloth and aligning the bottom edge of the cloth to a horizontal line; (2) Drop: dropping the cloth onto a flat surface and aligning its four corners to a flat square.
We compare four variants of our objective, each minimizes:
\begin{enumerate}
    \item $\mathcal{G}$ only: target error only.
    \item $\mathcal{G}+\mathcal{T}$: target error and trajectory irregularity.
    \item $\mathcal{G}+\mathcal{E}$: target error and potential energy.
    \item $\mathcal{G}+\mathcal{T} + \mathcal{E}$: target error and both regularization terms.
\end{enumerate}
Notice that the regularization terms $\mathcal{T}$ and $\mathcal{E}$ are scaled by $\alpha$ and $\beta$ respectively
to control their influences.
In the first task, we have $\alpha=1$ and $\beta = 1e-5$.
In the second task, we have $\alpha=1e-1$ and $\beta = 1e-5$.

\begin{figure}[!tb]
    \centering
    \includegraphics[width=0.95\linewidth]{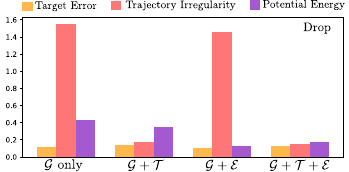}
    \vspace{-1em}
    \caption{Comparison of target error, trajectory irregularity and potential energy across four variants of our method
    on the drop task.
    From left to right, we optimize for
    $\mathcal{G}$ only, $\mathcal{G} + \mathcal{T}$, $\mathcal{G} + \mathcal{E}$ and $\mathcal{G} + \mathcal{T} + \mathcal{E}$.
    All cost values are padded by $0.1$ so some small values are visible in the same chart.}
    \label{fig:quantitative_result}
    \vspace{-1.5em}
\end{figure}

In Fig. \ref{fig:qualitative_result}, we show exemplar optimized cloth manipulation trajectories on two obstacle-free manipulation tasks.
Even though all variants of our method successfully reach the target, different combinations of regularization terms yield significantly different trajectories and cloth dynamics.
For $\mathcal{G}$ only, the cloth is moved to the target position but experiences a non-smooth trajectory and results in undesired wrinkles.
Optimizing $\mathcal{G} + \mathcal{T}$ together, the trajectory converges to a smooth straight line.
However, this ignores the dynamics of the cloth and causes extreme deformation.
Conversely, when optimizing for $\mathcal{G} + \mathcal{E}$, the cloth exhibits less deformation overall, but the trajectory can become inefficient and irregular.
The most favorable outcomes are generated when all three terms $\mathcal{G} + \mathcal{T} + \mathcal{E}$ are optimized simultaneously.
Fig. \ref{fig:quantitative_result} supports the same observation with quantitative data.
As aforementioned, the target error (yellow) is always minimized to a sufficiently small value.
However, by only optimizing $\mathcal{G} + \mathcal{T}$, the potential energy (purple) could still be large, which implies excessive deformation.
Meanwhile, if we only optimize $\mathcal{G} + \mathcal{E}$, trajectory irregularity (red) is high, resulting in inefficient trajectories.
It becomes evident that it is necessary to optimize all three terms $\mathcal{G} + \mathcal{T} + \mathcal{E}$ simultaneously
to achieve an accurate, efficient, and physically plausible manipulation trajectory.
\section{DISCUSSION \& CONCLUSIONS}

In this study, we introduce a cloth manipulation framework that places a strong emphasis on its awareness of cloth deformation physics and safety constraint satisfaction. Our framework leverages a differentiable XPBD simulation, which enables us to efficiently
minimize the cost function with quasi-Newton optimization method to discover the optimal control sequence. We confirm that the proposed framework is able to yield safe, physics-aware manipulation trajectories through a series of safety constraints experiments and ablation studies.
It holds the potential to advance robot autonomy in deformable object manipulation scenarios where safety and deformation mechanics are major concerns, such as autonomous surgery and elderly care.



\section*{ACKNOWLEDGMENT}
We sincerely appreciate Dr. Jie Fan of The BioRobotics Institute of Sant'Anna, Pisa, Italy, for her help during the preparation of this paper and experimental setups.

\label{Bibliography}
\bibliographystyle{unsrt} 
\bibliography{reference}

\end{document}